\documentclass{article} 
\usepackage{iclr2027_conference,times}

\usepackage{amsmath,amsfonts,bm}

\def\eqref#1{equation~\ref{#1}}

\def\1{\bm{1}}

\DeclareMathAlphabet{\mathsfit}{\encodingdefault}{\sfdefault}{m}{sl}
\SetMathAlphabet{\mathsfit}{bold}{\encodingdefault}{\sfdefault}{bx}{n}

\usepackage{graphicx}
\usepackage{hyperref}
\usepackage{url}
\usepackage[nameinlink,capitalise,noabbrev]{cleveref}
\definecolor{citationpink}{HTML}{D670C8}

\usepackage{booktabs}
\usepackage{multirow}
\usepackage{siunitx}
\usepackage{xcolor}
\usepackage{colortbl}
\usepackage{pifont}
\usepackage{listings}
\usepackage{tcolorbox}

\usepackage{tcolorbox}
\newtcolorbox{AIbox}[1]{
    title={#1},
    fonttitle=\bfseries,
    coltitle=black,
    colbacktitle=gray!15,
    colback=gray!3,
    colframe=gray!60,
    boxrule=0.5pt,
    arc=2pt,
    left=6pt,
    right=6pt,
    top=5pt,
    bottom=5pt
}
\lstdefinestyle{mobileprompt}{
    basicstyle=\ttfamily\footnotesize,
    columns=fullflexible,
    keepspaces=true,
    breaklines=true,
    breakatwhitespace=false,
    breakindent=1em,
    showstringspaces=false,
    numbers=none,
    aboveskip=4pt,
    belowskip=4pt
}

\hypersetup{
    colorlinks=true,
    citecolor=citationpink,
    linkcolor=black,
    urlcolor=black
}

\title{Learn How to Act from Your Own Interactions: On-Policy Self-Distillation for GUI Agents}

\newcommand{\authorline}[1]{%
  \makebox[\dimexpr\textwidth-2\tabcolsep\relax][c]{\normalfont #1}%
}

\author{%
\authorline{\textbf{%
Yan~Zhang\textsuperscript{1,3,5},
Daiqing~Wu\textsuperscript{4},
Huawen~Shen\textsuperscript{3},
Liang~Li\textsuperscript{1},
Gang~Cao\textsuperscript{3},%
}}\\[3pt]
\authorline{\textbf{%
Zhi~Gong\textsuperscript{3},
Wei~Dai\textsuperscript{3},
Xiaode~Zhang\textsuperscript{3},
Can~Ma\textsuperscript{1,$\dagger$},
Yu~Zhou\textsuperscript{2,$\dagger$}%
}}\\[5pt]
\authorline{\small
\textsuperscript{1}Institute of Information Engineering,
Chinese Academy of Sciences%
}\\[1pt]
\authorline{\small
\textsuperscript{2}VCIP \& TMCC \& DISSec,
College of Computer Science, Nankai University%
}\\[1pt]
\authorline{\small
\textsuperscript{3}Tencent
\qquad
\textsuperscript{4}Department of Psychological and Cognitive SciencesTsinghua University%
}\\[1pt]
\authorline{\small
\textsuperscript{5}School of Cyber Security,
University of Chinese Academy of Sciences%
}
}

\iclrfinalcopy 
\begin{document}

\maketitle

\begingroup
\renewcommand{\thefootnote}{\fnsymbol{footnote}}
\footnotetext[2]{Corresponding authors.}
\endgroup

\lhead{Under Review}

\begin{abstract}
Graphical User Interface (GUI) agents enable the fulfillment of complex user instructions through multi-turn interactions with software environments, requiring step-wise reasoning and long-horizon memory to guide actions and retain task-relevant information, respectively. Recent on-policy self-distillation (OPSD) methods have achieved strong performance on GUI grounding, a foundational subtask for GUI agents, owing to dense token-level supervision from privilege-conditioned self-teachers. However, extending existing OPSD methods to multi-turn GUI agents is hindered by self-teachers' limited privilege-following ability and insufficient privileged guidance. In this paper, we introduce GUI-SD-v2, the next version of GUI-SD, which extends OPSD from GUI grounding to multi-turn GUI interaction and addresses key limitations through a two-stage training framework. Specifically, GUI-SD-v2 first strengthens privilege following by jointly optimizing rollouts with and without privileged guidance from the same GUI states. Furthermore, it selectively distills step-specific reasoning and memory guidance through a privilege-conditioned self-teacher, supporting action decisions and the retention of task-relevant information for subsequent interactions. Extensive experiments on two representative GUI agent benchmarks, AndroidWorld and MobileWorld, show that GUI-SD-v2 compares favorably with existing OPSD baselines while consistently outperforming the evaluated state-of-the-art methods in both Pass@1 and Pass@3 success rates. Code and training data will be publicly released.

\end{abstract}

\section{Introduction}
Autonomous GUI agents have emerged as a promising direction in human--computer interaction, enabling applications ranging from personal assistance to large-scale software testing. In parallel, on-policy self-distillation (OPSD) offers an effective paradigm for post-training such agents, combining dense token-level supervision, informative learning signals for hard cases, and efficient single-rollout training. Within the GUI domain, GUI-SD \citep{zhang2026learn} presents the first exploration of OPSD for GUI grounding by conditioning the self-teacher on visually enriched privileged information and applying entropy-guided optimization to salient coordinate tokens. Subsequent work further refines GUI-SD in privileged information design and token-level optimization. Specifically, one line of work filters token-level distillation signals according to teacher confidence \citep{huang2026trust}, while another introduces diagnostic reflections and next-step GUI screenshots as alternative forms of privileged information \citep{xuan2026test, li2026next}.

\begin{figure}[t]
\begin{center}
\includegraphics[width=1.0\textwidth]{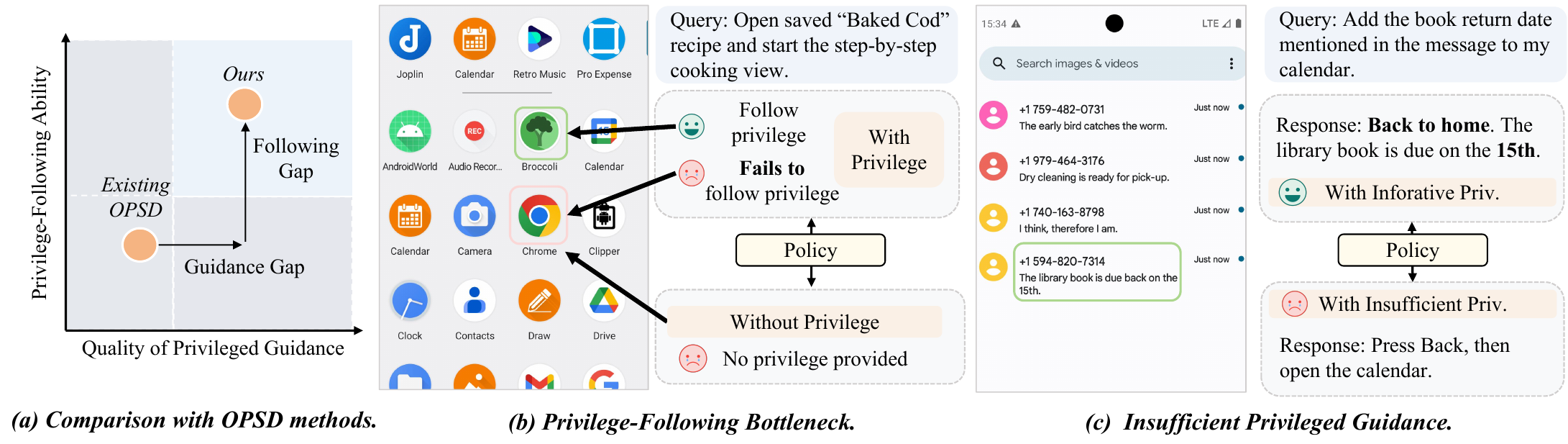}
\end{center}
\vspace{-8pt}
\caption{
\textbf{Motivation for GUI-SD-v2.}
(a) Limited privilege following prevents existing OPSD methods from reliably using privileged guidance for action decisions.
(b) Coarse privileges offer insufficient reasoning and memory guidance for multi-step interactions.
(c) GUI-SD-v2 improves both privilege following and privileged-guidance quality.
}
\label{fig:figure1}
\vspace{-8pt}
\end{figure}

Despite recent advances in GUI grounding, existing OPSD methods remain limited in both privilege-following ability and the quality of privileged guidance, hindering their extension to long-horizon GUI agents, as illustrated in \Cref{fig:figure1}(a): 1) \textbf{Privilege-Following Bottleneck.} Existing OPSD methods \citep{zhang2026learn,pei2026negative} suffer from limited privilege-following ability, which constrains the self-teacher's capacity to translate privileged guidance into reliable token-level supervision. As shown in \Cref{fig:figure1}(b), the policy still incorrectly select Chrome instead of the Broccoli recipe app even when privileged guidance is provided, indicating that the guidance does not reliably inform its action decisions. 2) Existing OPSD methods rely on coarse ground-truth cues or short self-generated reasoning traces \citep{li2026policy,yang2026matching}, offering limited guidance on action reasoning at the current step and retaining task-relevant information as interaction history for subsequent steps. As shown in \Cref{fig:figure1}(c), insufficient guidance focuses on navigating to the calendar without retaining the book's due date, which is essential information across steps.

To address these issues, we introduce GUI-SD-v2 (GUI Agent via Self-Distillation), the next version of GUI-SD that extends OPSD from GUI grounding to long-horizon GUI agent tasks. GUI-SD-v2 introduces privilege-following optimization to strengthen the policy's ability to follow privileged guidance, and informative privilege distillation to provide richer supervision across multi-step interactions. Specifically, privilege-following optimization constructs corrective guidance from interaction failures to augment the policy's training context. The policy is then optimized using a shared group of rollouts generated from the same GUI states with and without the guidance, strengthening privilege following for subsequent self-distillation stage. Subsequently, informative privilege distillation conditions the self-teacher on reasoning and memory privileges that guide action decisions and identify task-relevant information to retain as history, transferring these signals to the corresponding policy outputs through selective token-level self-distillation.


Extensive experiments on MobileWorld \citep{kong2026mobileworld} and AndroidWorld \citep{rawles2025androidworld} demonstrate that GUI-SD-v2 consistently outperforms existing OPSD methods and the evaluated state-of-the-art methods  under both Pass@1 and Pass@3. Detailed ablations further show that privilege-following optimization jointly improves privilege following and general GUI capabilities, while informative privilege distillation provides complementary token-level supervision for action reasoning and task-relevant memory throughout multi-step interactions.

Our main contributions are summarized as follows:
\begin{itemize}

\item Extending OPSD to long-horizon GUI agents reveals two key bottlenecks in existing methods, namely limited privilege following and insufficient guidance for reasoning and interaction memory.

\item We propose GUI-SD-v2, the next version of GUI-SD, which integrates privilege-following optimization with informative privilege distillation to strengthen privilege following and provide targeted reasoning and memory supervision across multi-step interactions.

\item Extensive experiments on MobileWorld and AndroidWorld demonstrate that GUI-SD-v2 consistently outperforms existing OPSD methods and evaluated state-of-the-art methods under both Pass@1 and Pass@3. 

\end{itemize}

\section{Related Work}
\subsection{On-Policy Self-Distillation for GUI Agents.}
OPSD \citep{li2026policy,lu2026self,zhang2026opsdl,sang2026policy,ye2026policy} has emerged as an effective post-training paradigm in which the same model serves as both student and teacher, with the teacher conditioned on privileged information to provide dense token-level supervision for student-generated on-policy responses. Within the GUI-agent domain, GUI-SD \citep{zhang2026learn} pioneers the application of OPSD to GUI grounding by conditioning the self-teacher on a visually enriched privileged context and applying entropy-guided optimization to salient coordinate tokens. Recent GUI-oriented OPSD methods \citep{huang2026trust,li2026next,xuan2026test,wu2026litegui,lian2026ui} mainly explore how to calibrate teacher signals and incorporate alternative forms of privileged information. Specifically, quality-aware self-distillation \citep{huang2026trust} combine correctness-aware gating with teacher-confidence scaling to filter unreliable coordinate-token signals and calibrate the remaining supervision. R-OPSD \citep{xuan2026test} derives diagnostic reflections from evaluated grounding attempts and conditions the self-teacher on these reflections to enable annotation-free test-time adaptation. However, current work remain centered on local decisions and lack privileged contexts that capture multi-step planning rationales and task-relevant interaction history.

\subsection{Long-Horizon GUI Agent Training.}
Long-horizon GUI agents complete complex user tasks through extended interactions with dynamic interfaces, requiring multi-step reasoning, action planning, task-state tracking, and error recovery. To develop these capabilities, recent work \citep{yang2025zerogui,gu2025mobile,shi2025mobilegui,lai2026computerrl,xu2026mobilerl,li2025efficient,lu2025ui} has explored end-to-end reinforcement learning directly from agent-environment interactions. Motivated by the effectiveness of GRPO in reasoning tasks, ARPO pioneers GRPO-based outcome-reward training for long-horizon GUI agents, augmenting policy optimization with a replay buffer of successful trajectories \citep{lu2025arpo}. WebAgent-R1 further develops outcome-reward training for web agents by integrating asynchronous online exploration with reasoning-oriented warm-up in an end-to-end multi-turn RL framework \citep{wei2025webagentr1}. Furthermore, subsequent work improves credit assignment by estimating task progress, evaluating intermediate decisions, and assigning rewards to verifiable milestones. For instance, ProgRM predicts task-completion progress at each interaction step and uses LCS-based self-annotation to construct dense progress rewards for online policy optimization \citep{zhang2025progrm}. Building on these advances, GUI-SD-v2 further extends credit assignment to the token level through dense OPSD supervision.

\section{Method}
We present GUI-SD-v2, the next version of GUI-SD that extends OPSD from GUI grounding to long-horizon GUI agent tasks. As shown in \Cref{fig:figure2}, GUI-SD-v2 consists of two training stages. (a) \textbf{Privilege-Following Optimization} (\Cref{fig:figure2} Top) integrates privilege-context and original rollouts into a shared group for relative-advantage optimization, improving both privilege following and general GUI capabilities. (b) \textbf{Informative Privilege Distillation} (\Cref{fig:figure2} Bottom) conditions the self-teacher on reasoning and memory privileges to provide informative token-level supervision throughout multi-turn interactions.

\subsection{Task Formulation}
\label{sec:task_formulation}
GUI-SD-v2 formulate long-horizon GUI interaction as a sequential decision process. At each interaction step \(t\), the policy \(\pi_\theta\) generates a structured response \(y_t\) conditioned on the user instruction \(q\), the current GUI screenshot \(I_t\), and the interaction history \(h_{0:t-1}\) covering all preceding steps:
\[
y_t\sim\pi_\theta\!\left(\cdot\mid q,I_t,h_{0:t-1}\right).
\]
As shown in the gray response block of \Cref{fig:figure2}(c), the structured response \(y_t\) contains four fields: \texttt{thought}, \texttt{action}, \texttt{tool\_call}, and \texttt{memory}. The \texttt{thought} field contains the current reasoning process, the \texttt{action} field describes the intended operation, and the \texttt{tool\_call} field specifies the corresponding GUI action. The \texttt{memory} field records task progress, instruction-relevant information, and potential interaction errors at the current step, forming \(h_t\) for subsequent interactions. The GUI agent repeats this process until the user instruction is completed or the maximum interaction horizon is reached. Detailed interaction templates are provided in the \Cref{app:template}.

\subsection{Privilege-Following Optimization}

\begin{figure}[t]
\begin{center}
\includegraphics[width=1.0\textwidth]{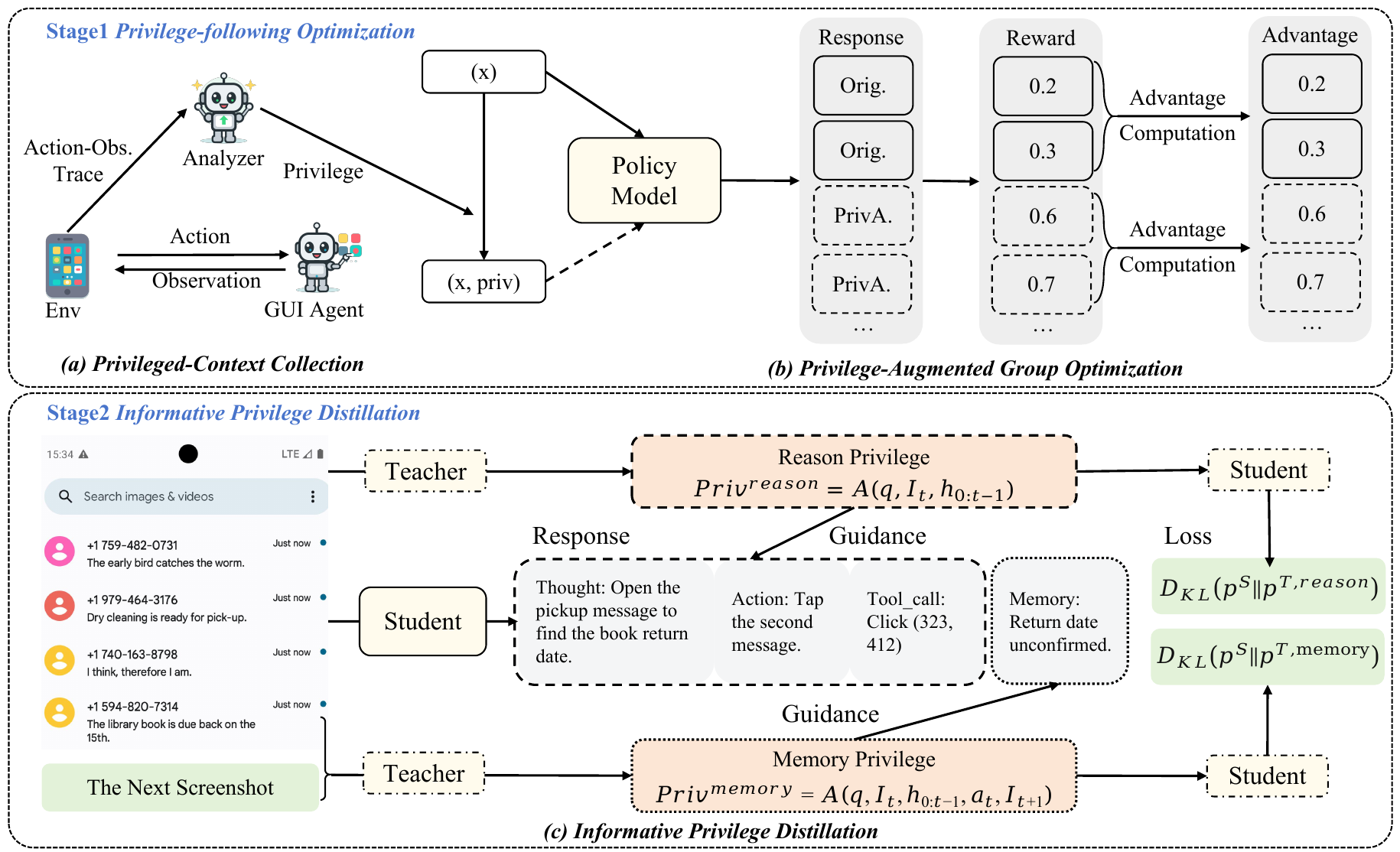}
\end{center}
\caption{
\textbf{Overview of GUI-SD-v2.}
\textbf{Stage~1 Privilege-Following Optimization.} (a) collects corrective privileges from GUI trajectories and (b) jointly optimizes original-context and privilege-augmented rollouts, strengthening privilege following for subsequent distillation. \textbf{Stage~2 Informative Privilege Distillation.} (c) transfers reasoning and memory guidance to decision and memory outputs, respectively, through self-distillation.
}
\vspace{-10pt}
\label{fig:figure2}
\end{figure}

To establish a reliable initialization for subsequent self-distillation, the first stage of GUI-SD-v2 (\Cref{fig:figure2} Top) equips the policy to follow the guidance provided by privileged contexts and generate the corresponding correct actions.

\textbf{Privileged-Context Collection.}
As illustrated in \Cref{fig:figure2}(a), the base policy generates GUI interaction trajectories for user instructions sampled from an offline pool with observations and actions recorded at each step. The analyzer model identifies an erroneous decision step in the trajectory and generates the corresponding correct action and a corrective experience annotation, which explains how the decision should be revised and serves as the privileged context. To validate these annotations, we apply rejection sampling to draw eight candidate actions from the privilege-conditioned GUI model and retain only samples for which at least one candidate matches the analyzer-provided correct action. Further details are provided in the \Cref{app:training_details}.

\textbf{Privilege-Augmented Policy Optimization.}
As illustrated in \Cref{fig:figure2}(b), given training samples with privileged annotations, we represent each sample as \((x,priv,a^{*})\), where \(x\) comprises the user instruction, current GUI screenshot, and interaction history from all preceding steps, \(priv\) denotes the privileged information, and \(a^{*}\) denotes the target action. For each sample, the policy constructs a shared group of \(N\) rollouts, comprising \(N/2\) original-context rollouts conditioned on \(x\) and \(N/2\) privilege-augmented rollouts conditioned on \((x,priv)\), all evaluated against \(a^{*}\).

Following prior work \citep{lu2026ui,luo2025gui}, GUI-SD-v2 assigns each rollout a verifiable reward based on response format and action correctness. The format component verifies the response structure defined in \cref{sec:task_formulation}, while the accuracy component evaluates the predicted GUI action against \(a^{*}\) using action-specific criteria, including target-region matching for grounding, semantic matching for text input, and exact action-type matching for functional operations. The two components are combined to obtain the final rollout reward.

To jointly improve privileged-context following and general GUI capabilities, we compute advantages by normalizing rewards independently within the privilege-augmented (\(c=\mathrm{priv}\)) and original-context (\(c=\mathrm{orig}\)) rollout subsets:
\[
A_i^{c}
=
\frac{R_i^{c}-\mu_c}{\sigma_c}.
\]
Here, \(i\) indexes a rollout, \(c\) identifies its subset, \(R_i^c\) is the corresponding reward, and \(\mu_c\) and \(\sigma_c\) are the reward mean and standard deviation within that subset. These subset-specific advantages are then jointly used to optimize the policy under the standard clipped GRPO objective:
\[
\mathcal{L}_{\mathrm{PAPO}}(\theta)
=
-\frac{1}{N}
\sum_{c\in\{\mathrm{priv},\mathrm{orig}\}}
\sum_{i=1}^{N/2}
\mathbb{E}_{t}\!\left[
\min\!\left(
\rho_{i,t}^{c}A_i^{c},
\operatorname{clip}\!\left(
\rho_{i,t}^{c},1-\delta,1+\delta
\right)A_i^{c}
\right)
\right].
\]
where \(\rho_{i,t}^c\) denotes the importance ratio between the updated and rollout policies, and \(\delta\) is the clipping threshold. The normalized advantage determines the contribution of each rollout, while ratio clipping constrains the magnitude of policy updates. Rollouts from both subsets jointly update the same policy, improving privileged-context following and general GUI capabilities.

\subsection{Informative Privilege Distillation}
As illustrated in \Cref{fig:figure2}(c), the second stage provides token-level supervision at each interaction step for reasoning over the user instruction, and for summarizing the current step action as task-relevant interaction history. Specifically, the current policy generates on-policy GUI trajectories, and the teacher simultaneously provides informative privileges at every interaction step of each trajectory. The policy is then trained through self-distillation to internalize the teacher-provided privileges, with its privilege-conditioned predictions supervising those under the original context.

\textbf{Informative Privilege Construction.}
For each user instruction, GUI-SD-v2 rolls out the current policy in the Android emulator environment to produce a multi-step GUI trajectory. Learning effectively from such long-horizon trajectories requires step-level supervision for two core capabilities: reasoning over the current interaction context to determine subsequent actions and summarizing each interaction step into task-relevant history for future decisions. Accordingly, the teacher generates complementary reasoning and memory privileges for every interaction step in the trajectory. At interaction step \(t\), the teacher analyzes the current GUI screenshot and accumulated interaction history in light of the user instruction to infer the current task state and reason about the next step. Its resulting think process serves as the reasoning privilege, providing step-specific guidance for subsequent reasoning and action selection. To construct the memory privilege, the teacher additionally observes the executed action and resulting next GUI state. By analyzing this state transition, it identifies task progress, instruction-relevant information, and salient state changes, with the resulting summary serving as the memory privilege.

\textbf{Informative Privilege-Guided OPSD.}
Each privilege provides targeted guidance that is incorporated into the policy through self-distillation. Specifically, for each on-policy response, the policy under the original context serves as the student, while the same policy under the corresponding privilege-augmented context serves as the self-teacher. Let \(p_j^{\mathrm{S}}\) denote the student distribution at token position \(j\), and let \(p_j^{\mathrm{T},\mathrm{reason}}\) and \(p_j^{\mathrm{T},\mathrm{memory}}\) denote the self-teacher distributions conditioned on the reasoning and memory privileges, respectively.

The reasoning privilege provides supervision for the complete decision process, covering the \texttt{thought}, \texttt{action}, and \texttt{tool\_call} fields. Denoting their token positions by \(\mathcal{M}_{\mathrm{reason}}\), the reasoning distillation loss is
\[
\mathcal{L}_{\mathrm{RKL}}^{\mathrm{reason}}
=
\frac{1}{|\mathcal{M}_{\mathrm{reason}}|}
\sum_{j\in\mathcal{M}_{\mathrm{reason}}}
D_{\mathrm{KL}}
\left(
p_j^{\mathrm{S}}
\,\middle\|\,
p_j^{\mathrm{T},\mathrm{reason}}
\right).
\]

The memory privilege instead supervises only the \texttt{memory} field. To isolate this guidance from the current action prediction, the student and self-teacher share the same response prefix through \texttt{tool\_call}, while the memory privilege is provided only to the self-teacher. The student generates the memory field without observing the next-state screenshot. Denoting the subsequent memory-token positions by $\mathcal{M}_{\mathrm{memory}}$, the memory distillation loss is
\[
\mathcal{L}_{\mathrm{RKL}}^{\mathrm{memory}}
=
\frac{1}{|\mathcal{M}_{\mathrm{memory}}|}
\sum_{j\in\mathcal{M}_{\mathrm{memory}}}
D_{\mathrm{KL}}
\left(
p_j^{\mathrm{S}}
\,\middle\|\,
p_j^{\mathrm{T},\mathrm{memory}}
\right).
\]

The two signals are combined as
\(\mathcal{L}_{\mathrm{priv}}
=
\lambda_{\mathrm{reason}}\mathcal{L}_{\mathrm{RKL}}^{\mathrm{reason}}
+
\lambda_{\mathrm{memory}}\mathcal{L}_{\mathrm{RKL}}^{\mathrm{memory}}\).
This selective distillation enables the policy to reason about the current action and retain task-relevant information for subsequent steps.

\section{Experiments}
\subsection{Experimental Setup}
\textbf{Implementation Details.} We conduct all experiments using Qwen3-VL-Instruct-8B \citep{bai2025qwen3} as the base model and perform multi-turn rollouts in Android emulator environments \citep{rawles2025androidworld} on selected tasks from OpenMobile \citep{cheng2026openmobile} and MobileForge \citep{liu2026mobileforge}. Specifically, the first stage involves 400 tasks, with Kimi-K3 \citep{team2026kimi} providing corrective guidance for erroneous decision steps. The second stage involves 1,013 tasks, with Kimi-K3 \citep{team2026kimi} providing privileged guidance on action planning and task-relevant memory. Further training details are provided in \Cref{app:training_details}.

\textbf{Baselines.} To ensure a fair comparison, we keep the training tasks fixed across GUI-SD-v2 and all baselines and employ the same Kimi-K3 model to generate the supervision required by each method. Specifically, Kimi-K3 \citep{team2026kimi} performs multi-turn rollouts on the selected tasks in the Android emulator, and its resulting GUI trajectories serve as ground-truth supervision. Consistent with prior work \citep{zhang2026learn}, we evaluate GUI-SD-v2 against four baselines: SFT, GRPO, Naive OPSD, and GUI-SD-v1. Both \textbf{SFT} and \textbf{GRPO} are trained with Kimi-K3 annotations as ground truth: SFT is supervised on complete teacher responses, including reasoning, actions, and task-relevant memory, whereas the action annotations serve as reward references for GRPO. \textbf{Naive OPSD} directly conditions the self-teacher on the ground-truth action as textual privileged information. \textbf{GUI-SD-v1} represents ground-truth click targets as visual privileges, including bounding boxes and Gaussian masks, while providing ground-truth targets for non-click actions as textual privileged information.

\textbf{Evaluation Benchmarks.} We evaluate GUI-SD-v2 on two representative mobile-agent benchmarks: AndroidWorld \citep{rawles2025androidworld} and MobileWorld \citep{kong2026mobileworld}. \textbf{AndroidWorld} provides a reproducible Android emulator environment with verifiable state-based evaluation. It comprises 116 tasks across 20 real-world applications, covering everyday activities such as note-taking, appointment scheduling, and messaging. \textbf{MobileWorld} is a more challenging dynamic benchmark comprising 201 tasks across 20 applications, with greater emphasis on long-horizon and cross-application tasks. Following prior work \citep{cheng2026openmobile,li2026next,liu2026mobileforge}, we report results on its GUI-only subset. To obtain more reliable performance estimates and mitigate the impact of environmental variability, we evaluate each task multiple times and report both pass@1 and pass@3.

\subsection{Main Result}

\begin{table}[t]
\centering

\setlength{\belowcaptionskip}{5pt}
\renewcommand{\arraystretch}{1.08}
\setlength{\tabcolsep}{6pt}
\setlength{\heavyrulewidth}{0.8pt}
\setlength{\lightrulewidth}{0.4pt}
\sisetup{table-format=2.1}

\caption{GUI Agent performance on two representative benchmarks, including AndroidWorld and MobileWorld. We report Pass@1 and Pass@3 success rates (\%). \textbf{Bold} indicates the best results.}

\begin{tabular}{@{}l S S S S@{}}
\toprule
\multirow{2}{*}{\textbf{Method}}
& \multicolumn{2}{c}{\textbf{AndroidWorld}}
& \multicolumn{2}{c}{\textbf{MobileWorld}} \\
\cmidrule(lr){2-3} \cmidrule(l){4-5}
& {Pass@1} & {Pass@3} & {Pass@1} & {Pass@3} \\
\midrule

GUI-Owl-7B \citep{xu2026mobile}
& 66.4 & {--} & {--} & {--} \\

DART-GUI-7B \citep{lu2026experience}
& 58.0 & {--} & {--} & {--} \\

UI-Venus-7B \citep{gu2025ui}
& 49.1 & {--} & 8.5 & {--} \\

UI-MOPD-8B \citep{lian2026ui}
& {--} & {--} & 12.0 & {--} \\

MobileForge-8B \citep{liu2026mobileforge}
& 50.9 & 67.2 & 10.3 & {--} \\

OpenMobile-8B \citep{cheng2026openmobile}
& 64.7 & 78.0 & 17.7 & 24.8 \\

GHD-8B \citep{li2026next}
& 66.5 & 73.3 & {--} & {--} \\

\midrule

Qwen3-VL-Instruct-8B \citep{bai2025qwen3}
& 40.5 & 55.2 & 17.9 & 23.9 \\

\quad \textit{+ SFT}
& 50.9 & 56.9 & 19.7 & 26.5 \\

\quad \textit{+ GRPO}
& 52.6 & 55.2 & 14.5 & 21.4 \\

\quad \textit{+ Naive OPSD}
& 50.0 & 53.4 & 13.7 & 19.7 \\

\quad \textit{+ GUI-SD-v1}
& 57.8 & 62.9 & 18.8 & 25.6 \\

\rowcolor{gray!15}
\quad \textbf{\textit{+ Ours}}
& \multicolumn{1}{c}{\textbf{67.2}\,{\scriptsize(+26.7)}}
& \multicolumn{1}{c}{\textbf{79.3}\,{\scriptsize(+24.1)}}
& \multicolumn{1}{c}{\textbf{25.6}\,{\scriptsize(+7.7)}}
& \multicolumn{1}{c@{}}{\textbf{31.6}\,{\scriptsize(+7.7)}} \\

\bottomrule
\end{tabular}

\label{tab:main_result}
\end{table}

\textbf{Comparisons with Baselines.}
\Cref{tab:main_result} reports Pass@1 and Pass@3 results on AndroidWorld \citep{rawles2025androidworld} and MobileWorld \citep{kong2026mobileworld}, two representative benchmarks for GUI Agents. GUI-SD-v2 achieves the highest success rates across both benchmarks and evaluation settings compared with baselines that share the same training-data scale and Kimi-K3 supervision source. Specifically, the \textit{SFT} baseline directly imitates Kimi-K3-generated responses, whereas the \textit{GRPO} baseline assigns rollout rewards based on step-level action correctness, taking Kimi-K3's action annotations as ground truth. Both \textit{SFT} and \textit{GRPO} rely on hard supervision based on Kimi-K3's reference responses and actions, with annotation errors potentially propagating into imitation targets and rollout rewards. Additionally, both \textit{Naive OPSD} and \textit{GUI-SD-v1} \citep{zhang2026learn} encode ground-truth actions as textual or visual privileges, offering limited guidance on action reasoning and interaction memory. GUI-SD-v2 combines step-specific reasoning and memory privileges with privilege-following optimization to provide richer supervision and strengthen the policy's ability to exploit it throughout multi-step interactions.

\textbf{Comparisons with SOTA Methods.}
GUI-SD-v2 also compares favorably with existing GUI-agent methods on AndroidWorld \citep{rawles2025androidworld} and MobileWorld \citep{kong2026mobileworld}, two representative benchmarks. On AndroidWorld, GUI-SD-v2 achieves Pass@1 and Pass@3 success rates of 67.2\% and 79.3\%, respectively, outperforming GHD \citep{li2026next}, which conditions the self-teacher on the next screenshot and applies distillation when the self-teacher corrects an erroneous student prediction. On MobileWorld, GUI-SD-v2 achieves 25.6\% Pass@1 and 31.6\% Pass@3, surpassing OpenMobile \citep{cheng2026openmobile} under both evaluation settings. Notably, GUI-SD-v2 obtains these gains by transferring Kimi-K3-generated reasoning and memory guidance into the GUI policy through OPSD-based training. This design provides targeted feedback on the policy's own interaction trajectories, supporting more reliable execution of long-horizon GUI tasks.

\subsection{Ablation Studies}

\begin{table}[t]
\centering
\renewcommand{\arraystretch}{1.15}
\setlength{\tabcolsep}{8pt}
\setlength{\belowcaptionskip}{8pt}

\caption{Comparison of training strategies for privilege-following optimization on AndroidWorld. PAPO denotes our proposed privilege-augmented policy optimization. Evaluation covers GUI capability after Stage~2, privilege-following accuracy after Stage~1, and GUI capability after Stage~1.}
\label{tab:priv_follow_ablation}

\begin{tabular}{lccc}
    \toprule
    \textbf{Method}
    & \shortstack{\textbf{GUI Capability}\\\textbf{(Stage 2)}}
    & \shortstack{\textbf{Privilege Following}\\\textbf{(Stage 1)}}
    & \shortstack{\textbf{GUI Capability}\\\textbf{(Stage 1)}} \\
    \midrule
    Baseline & 55.2 & 50.0 & 40.5 \\
    SFT      & 61.2 & 55.2 & 44.8 \\
    GRPO     & 60.3 & 57.8 & 44.8 \\
    \textbf{PAPO}
    & \textbf{67.2}
    & \textbf{62.9}
    & \textbf{49.1} \\
    \bottomrule
\end{tabular}

\end{table}

\textbf{Effectiveness of Privilege-Following Optimization.} Privilege-Following Optimization, the first stage of GUI-SD-v2, is designed to jointly improve the policy's privilege-following ability and general GUI capabilities. \Cref{tab:priv_follow_ablation} reports final AndroidWorld success rate after Stage~2 to assess whether this initialization benefits subsequent distillation, alongside task success rates after Stage~1 measured with and without privileged information provided at each rollout step to assess privilege following and general GUI capabilities, respectively.

Specifically, in row 1, the policy is trained directly in Stage~2 without privilege-following optimization, yielding the weakest results across all three evaluations and suggesting that limited privilege following constrains the effectiveness of informative privilege distillation. Row 2 adopts privilege-conditioned \textit{SFT} with annotated CoT responses as supervised targets, achieving 61.2\% AndroidWorld success after Stage~2. Row 3 instead optimizes privilege-conditioned rollouts through \textit{GRPO}, assigning rewards based on action correctness and achieving 60.3\% after Stage~2. Row 4 applies our proposed PAPO, jointly optimizing privilege-augmented and original-context rollouts within a shared group while normalizing their advantages separately. This design achieves 67.2\% final AndroidWorld success after Stage~2, together with 62.9\% privilege-following accuracy and 49.1\% general GUI performance after Stage~1, delivering the best results across all three evaluations. We attribute these gains to PAPO, which prepares the shared policy for both roles in Stage~2 by improving the self-teacher's use of privileged guidance and strengthening the student's general GUI capabilities.

\begin{table}[t]
\centering
\setlength{\belowcaptionskip}{8pt}
\renewcommand{\arraystretch}{1.12}
\setlength{\tabcolsep}{8pt}

\caption{Ablation of the memory and reasoning guidance components in informative privilege distillation. In our evaluation, the 116 AndroidWorld tasks are grouped into information retrieval (27 tasks) and task completion (89 tasks), with the former placing greater demands on retaining key information across interactions.}
\label{tab:reason_memory_ablation}

\begin{tabular}{ccccc}
\toprule
\textbf{Memory}
& \textbf{Reason}
& \textbf{AndroidWorld}
& \textbf{Information Retrieval}
& \textbf{Task Completion} \\
\midrule
\ding{55} & \ding{55} & 49.1
& 44.4~(12/27) & 50.6~(45/89) \\

\ding{55} & \ding{51} & 61.2
& 51.9~(14/27) & 64.0~(57/89) \\

\ding{51} & \ding{55} & 56.0
& 55.6~(15/27) & 56.2~(50/89) \\

\ding{51} & \ding{51} & \textbf{67.2}
& \textbf{59.3}~(16/27)
& \textbf{69.7}~(62/89) \\
\bottomrule
\end{tabular}
\end{table}

\begin{table}[t]
\centering
\renewcommand{\arraystretch}{1.08}
\setlength{\tabcolsep}{12pt}
\setlength{\belowcaptionskip}{8pt}

\caption{Effect of privileged-guidance granularity on AndroidWorld.
We report Pass@1 and Pass@3 success rates (\%).}
\label{tab:granularity}

\begin{tabular}{lcc}
    \toprule
    \multirow{2}{*}{\textbf{Granularity}}
    & \multicolumn{2}{c}{\textbf{AndroidWorld}} \\
    \cmidrule(lr){2-3}
    & \textbf{Pass@1} & \textbf{Pass@3} \\
    \midrule
    Trajectory-level & 51.7 & 60.3 \\
    \textbf{Step-level}& \textbf{67.2} & \textbf{79.3} \\
    \bottomrule
\end{tabular}
\vspace{-10pt}
\end{table}

\textbf{Effectiveness of Informative Privilege Distillation.}
\Cref{tab:reason_memory_ablation} evaluates informative privilege distillation by ablating its reasoning and memory guidance, reporting overall AndroidWorld success alongside performance on information-retrieval tasks requiring the retention of key information across interactions and task-completion tasks demanding multi-step reasoning. The first row reports a baseline without informative privilege distillation, in which neither reasoning nor memory guidance is provided, yielding the lowest success rates across all three evaluations. Additionally, introducing privileged reasoning guidance improves information-retrieval and task-completion success rates from 44.4\% to 51.9\% and from 50.6\% to 64.0\%, respectively, suggesting that better GUI-state interpretation and action planning benefit both task types. Incorporating privileged memory guidance increases information-retrieval success from 44.4\% to 55.6\%, suggesting that explicit memory supervision helps the policy remember task-relevant information across interactions. Combining both privileges yields the best results across all three evaluations, supporting the complementary roles of reasoning and memory guidance in long-horizon GUI interactions.

\textbf{Effect of Supervision Granularity.}
To investigate the effect of privileged-guidance granularity, we compare trajectory-level and step-level variants, both incorporating reasoning and memory guidance. The trajectory-level variant provides guidance for the entire trajectory and applies the same guidance at every interaction step, whereas our step-level variant provides guidance tailored to each step. As shown in \Cref{tab:granularity}, step-level supervision improves AndroidWorld Pass@1 from 51.7\% to 67.2\%, suggesting that matching reasoning and memory guidance to individual steps provides more effective supervision for long-horizon interactions.

\begin{table}[t]
\centering
\setlength{\belowcaptionskip}{8pt}
\renewcommand{\arraystretch}{1.08}
\setlength{\tabcolsep}{16pt}

\caption{Effect of privilege guidance from different teachers on AndroidWorld.}
\label{tab:teacher}

\begin{tabular}{lc}
    \toprule
    \textbf{Teacher} & \textbf{AndroidWorld} \\
    \midrule
    Qwen3.5-Plus \citep{qwen35blog}         & 57.8 \\
    MiniMax-M3 \citep{lai2026minimax}       & 59.5 \\
    Gemini3.8-Flash \citep{team2023gemini}  & 63.7 \\
    Kimi-K3 \citep{team2026kimi}            & \textbf{67.2} \\
    \bottomrule
\end{tabular}
\vspace{-5pt}
\end{table}

\textbf{Effect of Privilege Guidance from Different Teachers.} We compare four external models as sources of reasoning and memory privileges for GUI-SD-v2. As shown in \Cref{tab:teacher}, guidance from Qwen3.5-Plus \citep{qwen35blog}, MiniMax-M3 \citep{lai2026minimax}, and Gemini3.8-Flash \citep{team2023gemini} yields AndroidWorld success rates of 57.8\%, 59.5\%, and 63.7\%, respectively. GUI-SD-v2 performs best with Kimi-K3-generated guidance, reaching 67.2\%. These results highlight the influence of the privilege source on self-distillation effectiveness and support our choice of Kimi-K3 for privilege generation.

\begin{figure}[]
\begin{center}
\includegraphics[width=1\textwidth]{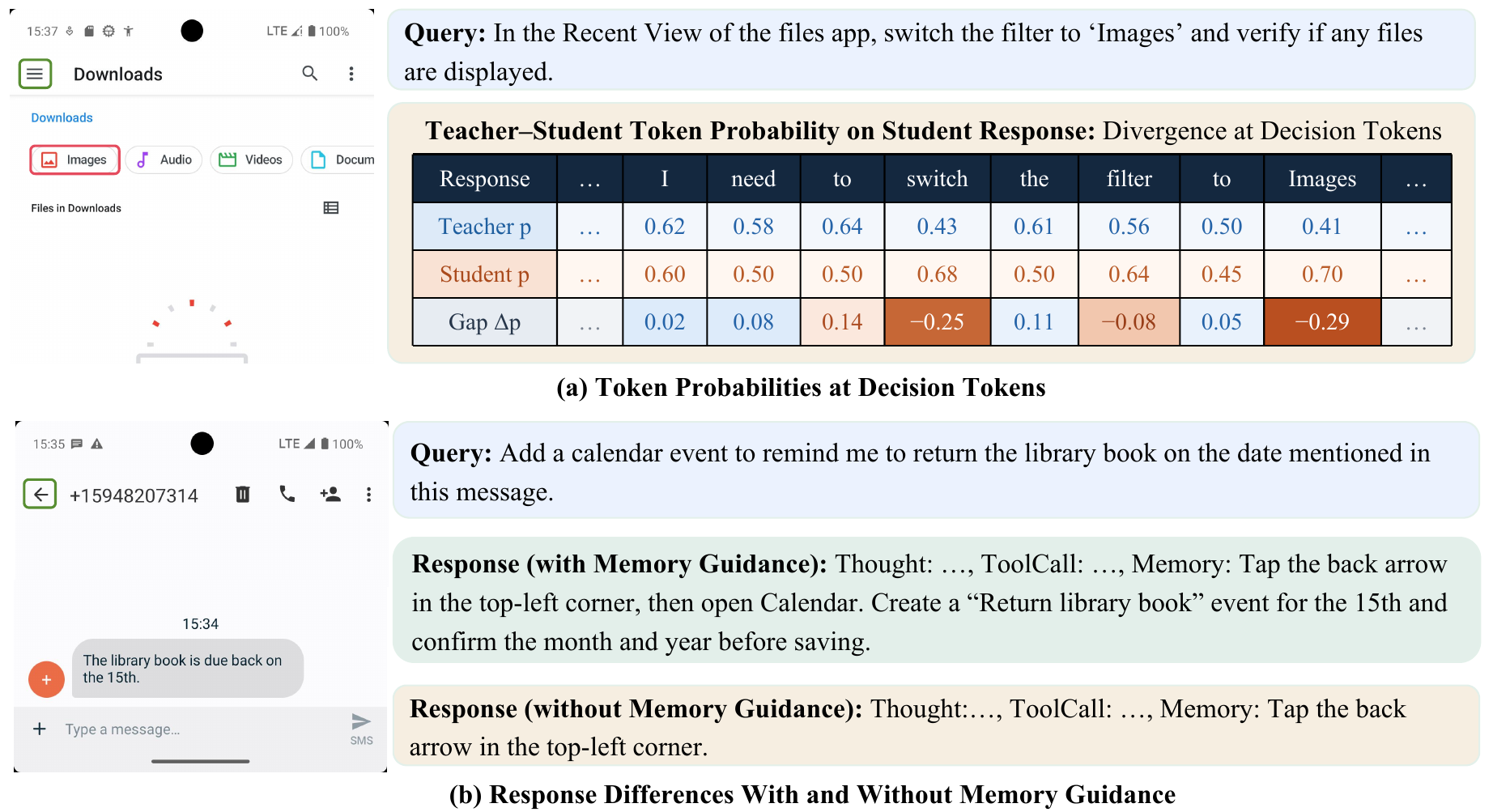}
\end{center}
\vspace{-10pt}
\caption{Qualitative analysis of informative privilege distillation.
(a) The self-teacher assigns lower probabilities to decision-critical tokens in an erroneous student response, providing a corrective distillation signal.
(b) Memory guidance encourages the policy to retain the task-relevant event details, whereas the unguided response retains only the navigation operation.}
\label{fig:figure3}
\vspace{-10pt}
\end{figure}

\subsection{Qualitative Analysis}
\textbf{Token Probabilities at Decision Tokens.}
As shown in \Cref{fig:figure3}(a), the self-teacher provides informative probability signals when its response is incorrect. When scoring the student's response, the teacher assigns lower probabilities to decision-critical tokens such as switch'' (0.43 versus 0.68) and Images'' (0.41 versus 0.70). These differences provide a distillation signal that discourages the erroneous action intent, illustrating the value of soft supervision beyond the correctness of the teacher's complete response.

\textbf{Response Differences With and Without Memory Guidance.}
As shown in \Cref{fig:figure3}(b), memory guidance encourages the policy to retain the phone number required by the instruction, whereas the response without memory guidance retains only the navigation operation. This comparison illustrates how memory supervision helps retain task-relevant information from the current step for subsequent decisions and task completion.

\section{Conclusion and Limitations}
We present GUI-SD-v2, a two-stage framework that extends on-policy self-distillation from GUI grounding to long-horizon GUI agents. By strengthening privilege following and selectively distilling step-specific reasoning and memory guidance, GUI-SD-v2 provides targeted token-level supervision across multi-turn interactions. Experiments on AndroidWorld and MobileWorld demonstrate consistent improvements over existing OPSD baselines and the evaluated state-of-the-art methods in both Pass@1 and Pass@3. While our evaluation focuses on Android environments, exploring additional GUI platforms and a broader range of model scales is a promising direction for future work.

\clearpage
\section*{Ethics Statement}
This work studies GUI-agent training using existing datasets and Android emulator environments. The main ethical considerations are shared with GUI-agent research more broadly, including user privacy, authorized access, and responsible automation. Deployment and reuse of the resulting models and data should respect applicable licenses and privacy requirements, operate within user-granted permissions, and maintain human oversight for consequential actions.

\section*{AI Use Statement}
We use Kimi-K3 to generate training supervision, including reference trajectories and responses, corrective action annotations, and step-specific reasoning and memory guidance, as described in the paper. Other generative models are compared as alternative sources of privileged guidance. AI tools also assist with manuscript drafting and editing. The authors take full responsibility for the research methodology, experimental results, and final manuscript.

\newpage

\bibliography{iclr2027_conference}

@article{lu2025arpo,
  title={Arpo: End-to-end policy optimization for gui agents with experience replay},
  author={Lu, Fanbin and Zhong, Zhisheng and Liu, Shu and Fu, Chi-Wing and Jia, Jiaya},
  journal={arXiv preprint arXiv:2505.16282},
  year={2025}
}

@inproceedings{wei2025webagentr1,
  title={Webagent-r1: Training web agents via end-to-end multi-turn reinforcement learning},
  author={Wei, Zhepei and Yao, Wenlin and Liu, Yao and Zhang, Weizhi and Lu, Qin and Qiu, Liang and Yu, Changlong and Xu, Puyang and Zhang, Chao and Yin, Bing and others},
  booktitle={Proceedings of the 2025 Conference on Empirical Methods in Natural Language Processing},
  pages={7920--7939},
  year={2025}
}

@article{li2026policy,
  title={On-Policy Self-Distillation without Any Supervision},
  author={Li, Yijiang and Wang, Bingyang and Liang, Yijun and Tian, Yunjie and Fu, Di and Vasconcelos, Nuno},
  journal={arXiv preprint arXiv:2608.06296},
  year={2026}
}

@article{lu2026self,
  title={Self-distilled agentic reinforcement learning},
  author={Lu, Zhengxi and Yao, Zhiyuan and Han, Zhuowen and Wang, Zi-Han and Wu, Jinyang and Gu, Qi and Cai, Xunliang and Lu, Weiming and Xiao, Jun and Zhuang, Yueting and others},
  journal={arXiv preprint arXiv:2605.15155},
  year={2026}
}

@article{zhang2026opsdl,
  title={Opsdl: On-policy self-distillation for long-context language models},
  author={Zhang, Xinsen and Ding, Zhenkai and Pan, Tianjun and Yang, Run and Kang, Chun and Xiong, Xue and Gu, Jingnan},
  journal={arXiv preprint arXiv:2604.17535},
  year={2026}
}

@article{sang2026policy,
  title={On-policy self-distillation for reasoning compression},
  author={Sang, Hejian and Xu, Yuanda and Zhou, Zhengze and He, Ran and Wang, Zhipeng and Sun, Jiachen},
  journal={arXiv e-prints},
  pages={arXiv--2603},
  year={2026}
}

@article{ye2026policy,
  title={On-policy context distillation for language models},
  author={Ye, Tianzhu and Dong, Li and Wu, Xun and Huang, Shaohan and Wei, Furu},
  journal={arXiv preprint arXiv:2602.12275},
  year={2026}
}

@article{zhang2026learn,
  title={Learn where to click from yourself: On-policy self-distillation for gui grounding},
  author={Zhang, Yan and Wu, Daiqing and Shen, Huawen and Ma, Can and Zhou, Yu},
  journal={arXiv preprint arXiv:2605.00642},
  year={2026}
}

@article{huang2026trust,
  title={Trust the Right Teacher: Quality-Aware Self-Distillation for GUI Grounding},
  author={Huang, Jingyuan and Huang, Zuming and Shi, Yucheng and Yang, Tianze and Zhai, Xiaoming and Chu, Wei and Liu, Ninghao},
  journal={arXiv preprint arXiv:2606.18101},
  year={2026}
}

@article{li2026next,
  title={The Next Screenshot Knows: Gated Hindsight Distillation for Mobile GUI Agents},
  author={Li, Weiwei and Liu, Junzhuo and Chu, Tong and Yu, Hengfu and Li, Wen},
  journal={arXiv preprint arXiv:2608.06065},
  year={2026}
}

@article{xuan2026test,
  title={Test-Time Self-Evolving GUI Visual Grounding via Reflection-Guided On-Policy Self-Distillation},
  author={Xuan, Shiyu and Li, Zechao},
  journal={arXiv preprint arXiv:2608.11191},
  year={2026}
}

@article{wu2026litegui,
  title={LiteGUI: Distilling Compact GUI Agents with Reinforcement Learning},
  author={Wu, Yubin and Cai, Zicheng and Ning, Liping and Wang, Hua and Chen, Zhi and Tang, Yaohua and Chen, Hao},
  journal={arXiv preprint arXiv:2605.07505},
  year={2026}
}

@article{lian2026ui,
  title={UI-MOPD: Multi-Platform On-Policy Distillation for Continual GUI Agent Learning},
  author={Lian, Niu and Chen, Alan and Yu, Zhehao and Duan, Chengzhen and Liu, Fazhan and Liu, Hui and Fu, Pei and Luan, Jian and Wang, Yaowei and Xia, Shu-Tao and others},
  journal={arXiv preprint arXiv:2607.04425},
  year={2026}
}

@article{yang2025zerogui,
  title={Zerogui: Automating online gui learning at zero human cost},
  author={Yang, Chenyu and Su, Shiqian and Liu, Shi and Dong, Xuan and Yu, Yue and Su, Weijie and Wang, Xuehui and Liu, Zhaoyang and Zhu, Jinguo and Li, Hao and others},
  journal={arXiv preprint arXiv:2505.23762},
  year={2025}
}

@article{gu2025mobile,
  title={Mobile-r1: Towards interactive reinforcement learning for vlm-based mobile agent via task-level rewards},
  author={Gu, Jihao and Ai, Qihang and Wang, Yingyao and Bu, Pi and Xing, Jingxuan and Zhu, Zekun and Jiang, Wei and Wang, Ziming and Zhao, Yingxiu and Zhang, Ming-Liang and others},
  journal={arXiv e-prints},
  pages={arXiv--2506},
  year={2025}
}

@article{shi2025mobilegui,
  title={Mobilegui-rl: Advancing mobile gui agent through reinforcement learning in online environment},
  author={Shi, Yucheng and Yu, Wenhao and Li, Zaitang and Wang, Yonglin and Zhang, Hongming and Liu, Ninghao and Mi, Haitao and Yu, Dong},
  journal={arXiv preprint arXiv:2507.05720},
  year={2025}
}

@inproceedings{lai2026computerrl,
  title={Computerrl: Scaling end-to-end online reinforcement learning for computer use agents},
  author={Lai, Hanyu and Liu, Xiao and Zhao, Yanxiao and Xu, Han and Zhang, Hanchen and Jing, Bohao and Ren, Yanyu and Yao, Shuntian and Dong, Yuxiao and Tang, Jie},
  booktitle={International Conference on Learning Representations},
  volume={2026},
  pages={23553--23591},
  year={2026}
}

@inproceedings{xu2026mobilerl,
  title={Mobilerl: Online agentic reinforcement learning for mobile gui agents},
  author={Xu, Yifan and Liu, Xiao and Liu, Xinghan and Fu, Jiaqi and Huang, Jiayu and Zhang, Hanchen and Jing, Bohao and Zhang, Shudan and Wang, Yuting and Dong, Yuxiao and others},
  booktitle={International Conference on Learning Representations},
  volume={2026},
  pages={35282--35315},
  year={2026}
}

@article{li2025efficient,
  title={Efficient multi-turn rl for gui agents via decoupled training and adaptive data curation},
  author={Li, Pengxiang and Hu, Zechen and Shang, Zirui and Wu, Jingrong and Liu, Yang and Liu, Hui and Gao, Zhi and Shi, Chenrui and Zhang, Bofei and Zhang, Zihao and others},
  journal={arXiv preprint arXiv:2509.23866},
  year={2025}
}

@article{lu2025ui,
  title={Ui-s1: Advancing gui automation via semi-online reinforcement learning},
  author={Lu, Zhengxi and Ye, Jiabo and Tang, Fei and Shen, Yongliang and Xu, Haiyang and Zheng, Ziwei and Lu, Weiming and Yan, Ming and Huang, Fei and Xiao, Jun and others},
  journal={arXiv preprint arXiv:2509.11543},
  year={2025}
}

@article{zhang2025progrm,
  title={Progrm: Build better gui agents with progress rewards},
  author={Zhang, Danyang and Zhang, Situo and Yang, Ziyue and Zhu, Zichen and Zhao, Zihan and Cao, Ruisheng and Chen, Lu and Yu, Kai},
  journal={arXiv preprint arXiv:2505.18121},
  year={2025}
}

@article{liu2026mobileforge,
  title={MobileForge: Annotation-Free Adaptation for Mobile GUI Agents with Hierarchical Feedback-Guided Policy Optimization},
  author={Liu, Guangyi and Zhao, Pengxiang and Wu, Gao and Yin, Yiwen and Li, Mading and Liu, Liang and Liu, Congxiao and Qi, Zhang and Wang, Mengyan and Guo, Liang and others},
  journal={arXiv preprint arXiv:2606.19930},
  year={2026}
}

@article{cheng2026openmobile,
  title={OpenMobile: Building open mobile agents with task and trajectory synthesis},
  author={Cheng, Kanzhi and Li, Zehao and Ma, Zheng and Chen, Nuo and Cao, Jialin and Sun, Qiushi and Ding, Zichen and Xu, Fangzhi and Yan, Hang and Chen, Jiajun and others},
  journal={arXiv preprint arXiv:2604.15093},
  year={2026}
}

@inproceedings{rawles2025androidworld,
  title={Androidworld: A dynamic benchmarking environment for autonomous agents},
  author={Rawles, Chris and Clinckemaillie, Sarah and Chang, Yifan and Waltz, Jonathan and Lau, Gabrielle and Fair, Marybeth and Li, Alice and Bishop, William and Li, Wei and Campbell-Ajala, Folawiyo and others},
  booktitle={International Conference on Learning Representations},
  volume={2025},
  pages={406--441},
  year={2025}
}

@inproceedings{kong2026mobileworld,
  title={Mobileworld: Benchmarking autonomous mobile agents in agent-user interactive and mcp-augmented environments},
  author={Kong, Quyu and Zhang, Xu and Yang, Zhenyu and Gao, Nolan and Liu, Chen and Tong, Panrong and Cai, Chenglin and Zhou, Hanzhang and Zhang, Jianan and Chen, Liangyu and others},
  booktitle={Proceedings of the 64th Annual Meeting of the Association for Computational Linguistics (Volume 1: Long Papers)},
  pages={6142--6167},
  year={2026}
}

@article{team2026kimi,
  title={Kimi k3: Open frontier intelligence},
  author={Team, Kimi and Bai, Tongtong and Bai, Yifan and Bao, Yiping and Cai, Jianfeng and Cai, Xinyuan and Cao, Peizhou and Cao, Yuxuan and Chai, Ziwei and Charles, Y and others},
  journal={arXiv preprint arXiv:2607.24653},
  year={2026}
}

@inproceedings{lu2026ui,
  title={Ui-r1: Enhancing efficient action prediction of gui agents by reinforcement learning},
  author={Lu, Zhengxi and Chai, Yuxiang and Guo, Yaxuan and Yin, Xi and Liu, Liang and Wang, Hao and Xiao, Han and Ren, Shuai and Zhao, Pengxiang and Liu, Guangyi and others},
  booktitle={Proceedings of the AAAI Conference on Artificial Intelligence},
  volume={40},
  number={21},
  pages={17608--17616},
  year={2026}
}

@article{luo2025gui,
  title={Gui-r1: A generalist r1-style vision-language action model for gui agents},
  author={Luo, Run and Wang, Lu and He, Wanwei and Chen, Longze and Li, Jiaming and Xia, Xiaobo},
  journal={arXiv preprint arXiv:2504.10458},
  year={2025}
}

@article{xu2026mobile,
  title={Mobile-agent-v3. 5: Multi-platform fundamental gui agents},
  author={Xu, Haiyang and Zhang, Xi and Liu, Haowei and Wang, Junyang and Zhu, Zhaozai and Zhou, Shengjie and Hu, Xuhao and Gao, Feiyu and Cao, Junjie and Wang, Zihua and others},
  journal={arXiv preprint arXiv:2602.16855},
  year={2026}
}

@inproceedings{lu2026experience,
  title={Experience-driven Multi-turn Reinforcement Learning for GUI Agents},
  author={Lu, Zhengxi and Ye, Jiabo and Tang, Fei and Shen, Yongliang and Xu, Haiyang and Zheng, Ziwei and Lu, Weiming and Yan, Ming and Huang, Fei and Xiao, Jun and others},
  booktitle={Proceedings of the 64th Annual Meeting of the Association for Computational Linguistics (Volume 1: Long Papers)},
  pages={9474--9496},
  year={2026}
}

@article{gu2025ui,
  title={Ui-venus technical report: Building high-performance ui agents with rft},
  author={Gu, Zhangxuan and Zeng, Zhengwen and Xu, Zhenyu and Zhou, Xingran and Shen, Shuheng and Liu, Yunfei and Zhou, Beitong and Meng, Changhua and Xia, Tianyu and Chen, Weizhi and others},
  journal={arXiv preprint arXiv:2508.10833},
  year={2025}
}

@article{bai2025qwen3,
  title={Qwen3-vl technical report},
  author={Bai, Shuai and Cai, Yuxuan and Chen, Ruizhe and Chen, Keqin and Chen, Xionghui and Cheng, Zesen and Deng, Lianghao and Ding, Wei and Gao, Chang and Ge, Chunjiang and others},
  journal={arXiv preprint arXiv:2511.21631},
  year={2025}
}

@misc{qwen35blog,
    title = {Qwen3.5: Towards Native Multimodal Agents},
    url = {https://qwen.ai/blog?id=qwen3.5},
    author = {Qwen Team},
    month = {February},
    year = {2026}
}

@article{lai2026minimax,
  title={Minimax sparse attention},
  author={Lai, Xunhao and Xu, Weiqi and Yang, Yufeng and Chen, Qiaorui and Xu, Yang and Zeng, Lunbin and Li, Xiaolong and Sun, Haohai and Zhu, Haichao and Zhang, Vito and others},
  journal={arXiv preprint arXiv:2606.13392},
  year={2026}
}

@article{team2023gemini,
  title={Gemini: a family of highly capable multimodal models},
  author={Team, Gemini and Anil, Rohan and Borgeaud, Sebastian and Alayrac, Jean-Baptiste and Yu, Jiahui and Soricut, Radu and Schalkwyk, Johan and Dai, Andrew M and Hauth, Anja and Millican, Katie and others},
  journal={arXiv preprint arXiv:2312.11805},
  year={2023}
}

@article{pei2026negative,
  title={Negative Self-Distillation: Learning to Reason by Avoiding Flaws},
  author={Pei, Rongcan and Wei, Zhepei and Xu, Shuyao and Zhu, Xinyu and Chen, Wei-Lin and Meng, Yu},
  journal={arXiv preprint arXiv:2609.11699},
  year={2026}
}

@article{yang2026matching,
  title={Matching Supervision to the Student's Learning Capacity: A Unified Framework for On-Policy Self-Distillation},
  author={Yang, Yongkang and Hao, Zhezheng and Zhang, Hong and Liu, Yi and Lin, Xiankun and Ji, Wence and Wei, Fanjunduo and Yu, Jiarui and Lin, Qiang and Liang, Xiaoyun and others},
  journal={arXiv preprint arXiv:2608.08176},
  year={2026}
}

@misc{zhao2024swift,
      title={SWIFT:A Scalable lightWeight Infrastructure for Fine-Tuning},
      author={Yuze Zhao and Jintao Huang and Jinghan Hu and Xingjun Wang and Yunlin Mao and Daoze Zhang and Zeyinzi Jiang and Zhikai Wu and Baole Ai and Ang Wang and Wenmeng Zhou and Yingda Chen},
      year={2024},
      eprint={2408.05517},
      archivePrefix={arXiv},
      primaryClass={cs.CL},
      url={https://arxiv.org/abs/2408.05517},
}
\bibliographystyle{iclr2027_conference}

\clearpage
\appendix
\crefalias{section}{appendix}

\section*{Appendix}

The appendix includes the following aspects:
\begin{itemize}
    \item \Cref{app:benchmarks}: Evaluation Benchmarks.
    \item \Cref{app:training_details}: Training Details.
    \item \Cref{app:eval_setting}: Evaluation Setting.
    \item \Cref{app:template}: GUI-SD-v2 Template.
\end{itemize}
    
\section{Evaluation Benchmarks}
\label{app:benchmarks}
\textbf{AndroidWorld.}
AndroidWorld \citep{rawles2025androidworld} is an interactive benchmark for evaluating autonomous agents in a fully functional Android environment. It comprises 116 parameterized task templates across 20 applications, covering everyday activities such as messaging, calendar management, note-taking, file manipulation, and system configuration. The benchmark includes tasks that modify application or device states and information-retrieval tasks that require answering questions about existing data. Each task is dynamically instantiated with randomized parameters and dedicated initialization, validation, and teardown procedures, supporting reproducible evaluation under varying task conditions. Task outcomes are assessed programmatically by inspecting application databases, files, and system settings, or by checking the agent's answer against the expected response. This outcome-based evaluation accommodates different valid interaction sequences rather than requiring agents to reproduce a predefined action trajectory.

\textbf{MobileWorld.}
MobileWorld \citep{kong2026mobileworld} comprises 201 tasks across 20 applications, emphasizing long-horizon workflows and cross-application interactions in domains such as communication, e-commerce, productivity, and information retrieval. The full benchmark contains 117 GUI-only tasks, 45 agent-user interaction tasks requiring clarification of incomplete instructions, and 40 tasks combining GUI operations with Model Context Protocol (MCP) tools. Across the complete benchmark, 62.2\% of tasks involve multiple applications, requiring agents to coordinate actions and maintain relevant information across application boundaries. The environment supports reproducible evaluation through snapshot-based initialization and self-hosted application backends. Task success is determined through textual answer verification, backend database checks, local storage inspection, or application-specific callbacks. In this work, we evaluate on the GUI-only subset, which includes both task-completion and information-retrieval tasks, focusing on multi-step GUI interaction without requiring user clarification or MCP tool use.

\section{Training Details}
\label{app:training_details}

\subsection{Hyperparameters}

We implement training using the ms-swift framework
\citep{zhao2024swift} with Qwen3-VL-8B-Instruct
\citep{bai2025qwen3} as the base model.
All model parameters, including the vision encoder and
vision-language connector, are optimized using AdamW.
\Cref{tab:training_hyperparameters} summarizes the hyperparameters
for both stages.

\begin{table}[t]
\centering
\setlength{\belowcaptionskip}{8pt}
\renewcommand{\arraystretch}{1.08}
\setlength{\tabcolsep}{12pt}

\caption{Training hyperparameters for the two stages of GUI-SD-v2.}
\label{tab:training_hyperparameters}

\begin{tabular}{@{}ll@{}}
\toprule
\textbf{Hyperparameter} & \textbf{Value} \\
\midrule
\multicolumn{2}{l}{\textbf{Stage 1: Privilege-Following Optimization}} \\
\midrule
Learning rate & $5\times10^{-6}$ \\
Learning-rate schedule & Cosine decay to zero \\
Warmup ratio & 0.05 \\
Training epochs & 1 \\
Per-device micro-batch size & 2 \\
Gradient accumulation steps & 16 \\
Rollouts per GUI state & 8 \\
Rollouts with / without privilege & 4 / 4 \\
Sampling temperature & 1.0 \\
Top-$p$ & 1.0 \\
Max. response length (tokens) & 16,384 \\
\midrule
\multicolumn{2}{l}{\textbf{Stage 2: Informative Privilege Distillation}} \\
\midrule
Initialization & Stage~1 checkpoint \\
Learning rate & $1\times10^{-6}$ \\
Learning-rate schedule & Cosine decay to zero \\
Warmup ratio & 0.1 \\
Training epochs & 2 \\
Gradient accumulation steps & 16 \\
On-policy trajectories per task & 1 \\
Sampling temperature & 1.0 \\
Top-$p$ & 1.0 \\
Max. response length (tokens) & 16,384 \\
Max. trajectory length (steps) & 30 \\
\bottomrule
\end{tabular}
\end{table}

\subsection{Reward Specification}

Stage~1 evaluates generated responses using a format reward and
an action-specific accuracy reward.

\textbf{Format Reward.}
The format reward $R_{\mathrm{format}}$ checks whether the response
follows the required structure containing \texttt{thought},
\texttt{action}, \texttt{tool\_call}, and \texttt{memory}.

\textbf{Accuracy Reward.}
The accuracy reward $R_{\mathrm{accuracy}}$ is selected according
to the action category, using $R_{\mathrm{click}}$ for click actions,
$R_{\mathrm{text}}$ for text-input actions, and $R_{\mathrm{func}}$
for functional actions:
\begin{equation}
\begin{aligned}
R_{\mathrm{click}}
&=
\begin{cases}
1, & \|o_{\mathrm{point}}-p_{\mathrm{gt}}\|_2 \leq 50, \\
0, & \text{otherwise},
\end{cases}
\\[4pt]
R_{\mathrm{text}}
&=
\begin{cases}
1, & \mathrm{F1}(o_{\mathrm{text}},T_{\mathrm{gt}})>0.5, \\
0, & \text{otherwise},
\end{cases}
\\[4pt]
R_{\mathrm{func}}
&=
\begin{cases}
1, & o_{\mathrm{act}}=A_{\mathrm{gt}}, \\
0, & \text{otherwise}.
\end{cases}
\end{aligned}
\label{eq:action_accuracy_reward}
\end{equation}
Here, $o_{\mathrm{point}}$ and $p_{\mathrm{gt}}$ denote the predicted
and reference click coordinates, with Euclidean distance measured
in pixels. $\mathrm{F1}(\cdot)$ measures the agreement between the
predicted text $o_{\mathrm{text}}$ and reference text $T_{\mathrm{gt}}$.
For functional actions, $o_{\mathrm{act}}$ and $A_{\mathrm{gt}}$
denote the predicted and reference action types.

\textbf{Response Reward.}
The final reward combines format compliance and action accuracy:
\begin{equation}
R = 0.2R_{\mathrm{format}} + 0.8R_{\mathrm{accuracy}}.
\label{eq:response_reward}
\end{equation}

\subsection{Privileged-Context Collection}
GUI-SD-v2 collects trajectories by rolling out the base policy on 400 tasks from MobileForge \citep{liu2026mobileforge}, recording the observations and actions at each interaction step. Kimi-K3 \citep{team2026kimi} analyzes these trajectories to identify erroneous decisions and generate the corresponding reference actions and corrective experience annotations. Each annotation explains how the erroneous decision should be revised and serves as the privileged context for that step.

GUI-SD-v2 validates the annotations through rejection sampling before Stage~1 training. For each annotated step, the GUI policy receives the original context augmented with the corrective annotation and generates eight candidate responses. The sample is retained only if at least one candidate predicts an action matching the reference. The retained samples provide the GUI states, privileged contexts, and reference actions used for privilege-following optimization.

\section{Evaluation Settings}
\label{app:eval_setting}
\textbf{Evaluation Metrics.} GUI-SD-v2 is evaluated on each benchmark over three runs, with every task evaluated once per run and task seeds held fixed across runs. Pass\@1 is the mean success rate across the three complete benchmark runs, while Pass\@3 is the percentage of tasks successfully completed in at least one of the three attempts. Task success is determined by each benchmark's evaluator. Attempts affected by environment initialization failures are excluded from scoring and rerun, ensuring three valid attempts per task.

\textbf{Baseline Comparison.} All baselines use training data from both stages of GUI-SD-v2, sharing the same training-task pools and supervision source, Kimi-K3 \citep{team2026kimi}. The supervision format is adapted to each method: SFT learns from teacher-generated responses, while GRPO uses teacher-provided actions to compute step-level rewards. Naive OPSD and GUI-SD-v1 use reference actions as textual or visual privileges for self-distillation.

\section{GUI-SD-v2 Template}
\label{app:template}

The tool-definition template in \Cref{fig:prompt-mobile} specifies the available GUI actions, their required arguments, and the coordinate conventions for spatial operations. The response template in \Cref{fig:prompt-response} organizes each response into four fields: thought explains the next operation, action describes the intended action, tool call specifies the function invocation and its arguments, and memory captures task-relevant information to retain across interactions.

\begin{figure*}[p]
\begin{AIbox}{Mobile Interaction Prompt: Tool Definition}
\small

You are a mobile GUI assistant. Use the user instruction,
current screenshot, and interaction history to select the
next operation and retain information needed later.

\medskip
\textbf{\# Tools}

Coordinates are normalized to 0--999 on each axis,
with the origin at the top-left corner. Target the center
of the intended element and allow the interface to update
before choosing the next operation.

\begin{lstlisting}[style=mobileprompt]
<tools>
{
  "type": "function",
  "function": {
    "name": "mobile_use",
    "description": "Operate a mobile touchscreen and receive screenshots.",
    "parameters": {
      "type": "object",
      "properties": {
        "action": {
          "type": "string",
          "enum": ["click", "long_press", "swipe", "type",
                   "answer", "system_button", "wait", "terminate"]
        },
        "coordinate": {
          "type": "array",
          "description": "[x,y]: target for click/long_press; start for swipe."
        },
        "coordinate2": {
          "type": "array",
          "description": "[x,y]: ending point for swipe."
        },
        "text": {
          "type": "string",
          "description": "Input content for type; response content for answer."
        },
        "time": {
          "type": "number",
          "description": "Duration in seconds for long_press or wait."
        },
        "button": {
          "type": "string",
          "enum": ["Back", "Home", "Menu", "Enter"],
          "description": "System button selected by system_button."
        },
        "status": {
          "type": "string",
          "enum": ["success", "failure"],
          "description": "Task outcome reported by terminate."
        }
      },
      "required": ["action"]
    }
  }
}
</tools>
\end{lstlisting}

Include the arguments associated with the selected action.
For \texttt{swipe}, provide both coordinate fields.
For \texttt{system\_button}, use \texttt{Back} for the
previous screen, \texttt{Home} for the home screen,
\texttt{Menu} for recent apps, or \texttt{Enter} for
the Enter key.

\end{AIbox}
\caption{Tool definition for GUI-SD-v2 .}
\label{fig:prompt-mobile}
\end{figure*}

\begin{figure*}[p]
\begin{AIbox}{Mobile Interaction Prompt: Response Format}
\small

\textbf{\# Response Format}

Return exactly four parts in the following order.

\medskip
\noindent
\textbf{1. Thought.}
Within \texttt{<thought>...</thought>}, briefly explain
the next operation based on the current observation,
user instruction, and interaction history.

\medskip
\noindent
\textbf{2. Action.}
Within \texttt{<action>...</action>}, describe the
intended operation in one short sentence.

\medskip
\noindent
\textbf{3. Tool call.}
Within \texttt{<tool\_call>...</tool\_call>}, provide
exactly one valid JSON object. Set \texttt{name} to
\texttt{"mobile\_use"} and include the selected action
and its associated parameters in \texttt{arguments}.
The call must implement the operation described
in \texttt{action}.

\medskip
\noindent
\textbf{4. Memory.}
Within \texttt{<memory>...</memory>}, retain task-relevant facts from the current screenshot and interaction history, preserving relevant names, dates, numbers, and other details needed later. Summarize confirmed progress and observed errors. Record the intended action and any outcomes that require verification in subsequent steps. Update unresolved outcomes only when supported by new observations.

\medskip
\textbf{\# Rules}

Use \texttt{answer} with \texttt{text} to deliver
requested information. Use \texttt{terminate} with
\texttt{status} to end the task. Output no content
outside the four required parts.

\medskip
\textbf{\# Example}

The following illustrates a waiting step when the current
screen displays a loading indicator.

\begin{lstlisting}[style=mobileprompt]
<thought>
The loading indicator suggests that the interface needs more time.
</thought>
<action>
Wait one second for the interface to update.
</action>
<tool_call>
{"name": "mobile_use", "arguments": {"action": "wait", "time": 1}}
</tool_call>
<memory>
The current screen shows a loading indicator; completion is unverified.
</memory>
\end{lstlisting}

\end{AIbox}
\caption{Response instructions specifying a structured output with thought, action, tool call, and memory fields.}
\label{fig:prompt-mobile-format}
\end{figure*}

\end{document}